\documentclass[%
reprint,
amsmath,amssymb,
aps,
pra,
]{revtex4-2}

\usepackage{graphicx}
\usepackage{dcolumn}
\usepackage{bm}
\usepackage{braket}
\usepackage{amsmath}
\usepackage{amsfonts}
\usepackage{multirow}
 \usepackage{booktabs}

\begin{document}
	
	\preprint{APS/123-QED}
	
	\title{Enhancing Quantum Federated Learning with Fisher Information-Based Optimization}
	

 
	\author{Amandeep Singh Bhatia}
	\email{drasinghbhatia@gmail.com, asbhatia@ncsu.edu}
	
	\author{Sabre Kais}%
	\email{Corresponding author: skais@ncsu.edu}
	\affiliation{ Department of Electrical and Computer Engineering, North Carolina State University, Raleigh, NC, USA}
	
	\begin{abstract}
		       Federated Learning (FL) has become increasingly popular across different sectors, offering a way for clients to work together to train a global model without sharing sensitive data. It involves multiple rounds of communication between the global model and participating clients, which introduces several challenges like high communication costs, heterogeneous client data, prolonged processing times, and increased vulnerability to privacy threats. In recent years, the convergence of federated learning and parameterized quantum circuits has sparked significant research interest, with promising implications for fields such as healthcare and finance. By enabling decentralized training of quantum models, it allows clients or institutions to collaboratively enhance model performance and outcomes while preserving data privacy. Recognizing that Fisher information can quantify the amount of information that a quantum state carries under parameter changes, thereby providing insight into its geometric and statistical properties. We intend to leverage this property to address the aforementioned challenges. In this work, we propose a Quantum Federated Learning (QFL) algorithm that makes use of the Fisher information computed on local client models, with data distributed across heterogeneous partitions. This approach identifies the critical parameters that significantly influence the quantum model's performance, ensuring they are preserved during the aggregation process. Our research assessed the effectiveness and feasibility of QFL by comparing its performance against other variants, and exploring the benefits of incorporating Fisher information in QFL settings. Experimental results on ADNI and MNIST datasets demonstrate the effectiveness of our approach in achieving better performance and robustness against the quantum federated averaging method.

 
	\end{abstract}

	\maketitle
	\section{Introduction}

Protecting privacy has become essential in the current digital environment, especially within classical or quantum or hybrid machine learning applications. In recent years, Federated Learning (FL) has gained prominence as a distributed learning framework, enabling multiple clients (such as mobile devices or institutions) to collaboratively train models while maintaining decentralized data \cite{1}. This approach allows clients to share model parameters with a central server without revealing their private data. FL enhances data privacy and security, making it especially suitable for sensitive domains like healthcare and finance. In real-world scenarios, implementing Federated Learning (FL) presents distinct challenges, including data heterogeneity due to non-independent and non-identically distributed (non-IID) data, as well as substantial communication overhead \cite{2,3,4}. To address this, various strategies such as model compression and aggregation have been developed, proving to be highly impactful \cite{5,6}. Additionally, some federated learning approaches also leverage Fisher information for aggregation \cite{7,8}, showcasing its potential to improve model performance.

\begin{figure}[!ht]
	\centering
\includegraphics[scale=0.65]{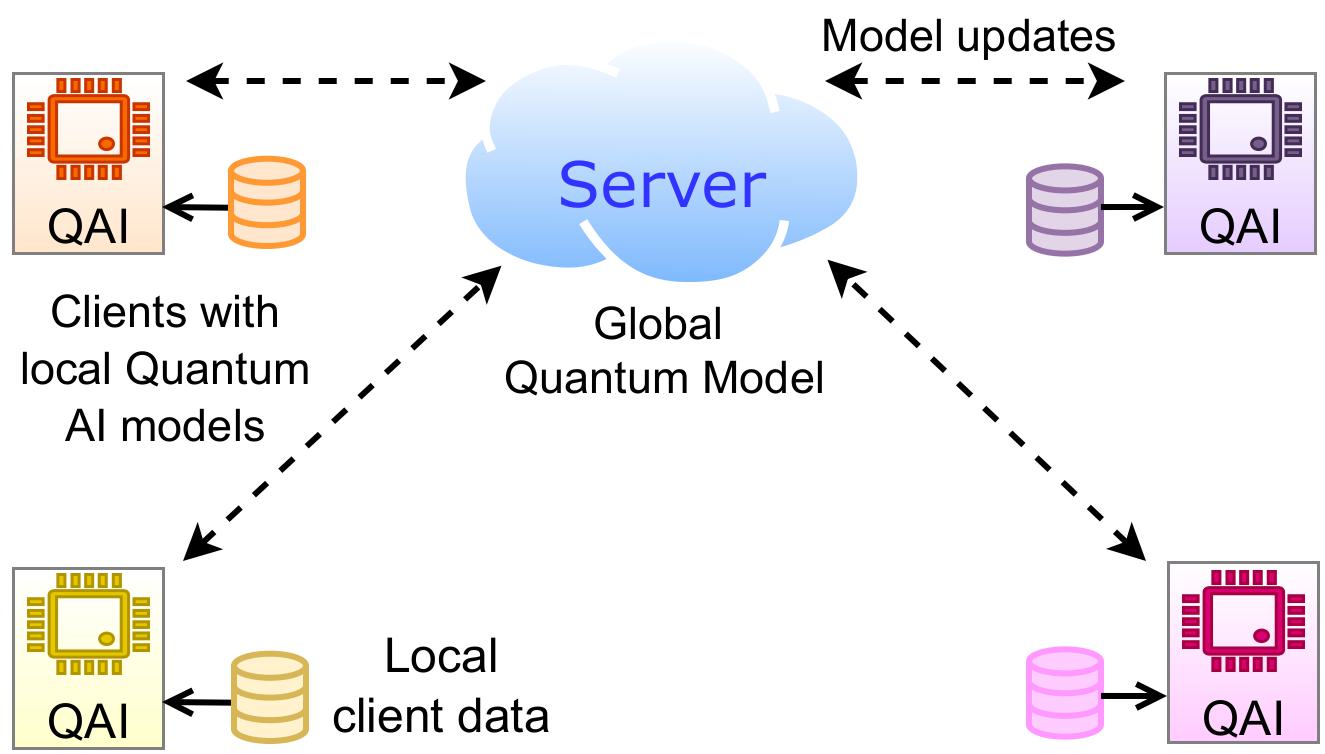}
\caption{\textbf{An illustration of the quantum federated learning process}. The central server initializes the global model with random weights and distributes them to clients (e.g., hospitals or edge devices). Clients train their local quantum AI models on their data and send updates to the server, which refines the global model through techniques like weighted averaging. This iterative process ensures data privacy by keeping raw data decentralized while collaboratively improving the global model until convergence.}
\label{qai}
\end{figure}

In parallel, quantum computing, especially quantum machine learning (QML) \cite{9, 10}, including quantum federated learning has made remarkable progress. Its ability to leverage the combined power of distributed quantum resources surpasses the constraints of individual quantum nodes. In recent years, Quantum Federated Learning (QFL) has become a rapidly evolving area of interest for academic and industrial communities. It has demonstrated the potential to significantly impact a wide range of practical applications across diverse domains, including healthcare, manufacturing, and finance. Noteworthy implementations of QML algorithms in federated settings include quantum neural networks \cite{11}, variational quantum tensor networks (QTNs) \cite{12}, variational quantum circuits (VQCs) \cite{14, 15, 100}, quanvolutional neural network \cite{16}, quantum convolutional neural networks (QCNNs) \cite{17}, quantum long short-term memory (QLSTM) \cite{18}, and federated QML for advancing drug discovery and healthcare applications by enabling collaborative data analysis while preserving privacy \cite{13}. Some studies significantly reduced data consumption, some reduced communication rounds, and others reduced the trainable parameters in FL settings.

\begin{figure*}[!ht]
	\centering
\includegraphics[scale=0.65]{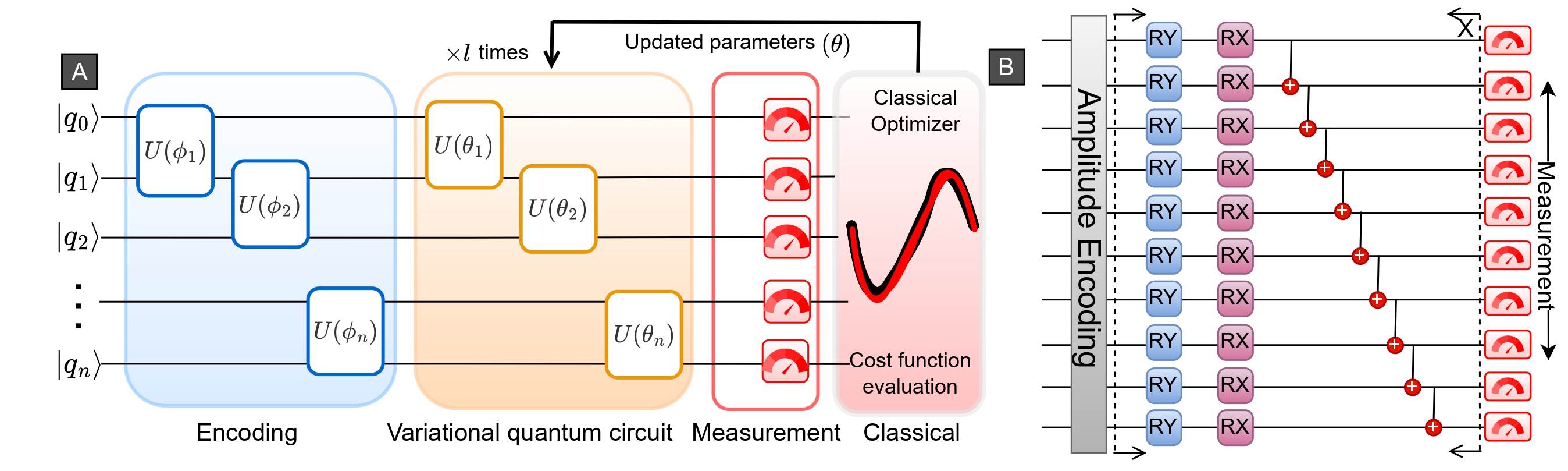}
\caption{\textbf{An overview of a variational quantum classifier (VQC)}. (A) It includes a variational quantum circuit consisting of a feature map that encodes classical data into a quantum state, a parameterized quantum circuit that applies quantum operations optimized during training, and quantum measurements performed on the transformed state. These measurements produce results corresponding to class probabilities or predictions. The output from the quantum circuit feeds into a classical optimization loop, which adjusts the circuit parameters using a predefined cost function. (B) The VQC  begins with a data-loading layer that amplitude-encodes the input image data into a quantum state. The variational component consists of multiple repeated layers, with only a single layer depicted for clarity. after encoding, the quantum state is transformed through a series of layers featuring trainable single-qubit rotations (RY and RX) and entangling CNOT gates. Finally, a measurement is performed to determine the classification label.}
\label{vqcfig}
\end{figure*}

In QFL, the federated averaging (FedAvg) performs well with independent and identically distributed (IID) data but struggles to converge in real-world non-IID scenarios. 
Variations in data distributions across clients hinder the model's ability to generalize, posing a significant challenge for the FL environment \cite{19, 20}.
 It aggregates the local gradients from all clients in the federation. 
While clients can independently train models suited to their local datasets, the server faces the difficulty of merging these into a unified global model that generalizes across all clients \cite{21}. 
QFL on non-IID distributions struggles to achieve high accuracy and requires more communication rounds compared to scenarios with IID data \cite{16, 17}.

\subsection{Motivation}
Evaluating quantum Fisher information typically requires full state tomography, which scales exponentially with qubit count, making it impractical for near-term applications. Variational quantum circuits offer a practical alternative due to their versatility and efficiency, while classical Fisher information quantifies a quantum model's sensitivity to parameter changes. By leveraging Fisher information matrices from local quantum models, our method captures variations in data distributions, effectively integrating diverse client datasets into a unified global model. To enhance quantum federated learning, we incorporate variational quantum circuits and utilize layer-wise Fisher information for improved aggregation.

\subsection{Main contributions}

To summarize, this paper makes the following contributions:
\begin{itemize}
    \item Proposal of a quantum federated learning framework based on Fisher information-based optimization, or QFedFisher, for collaborative learning between multiple clients, enabling QML models to train on non-independent and identically distributed data.
    \item We retained the useful parameters of each client using the layer-wise Fisher information from the quantum circuit. 
    By doing so, we protected these valuable parameters from being overwritten by potentially noisy or less significant global parameters during the aggregation, thereby maintaining the integrity and effectiveness of the local client contributions in the global model.
    \item As demonstrated by our experiments, the QFedFisher algorithm effectively preserves the key parameters of each client, leading to improved convergence and better performance compared to existing state-of-the-art QFL methods.
\end{itemize}

\section{Quantum federated learning}
In this section, we will explore how to leverage quantum federated learning by combining variational quantum circuits and classical Fisher information.  Our aim is to enable collaborative quantum training that enhances accuracy and privacy in federated learning settings, utilizing Fisher information matrices computed at local quantum models. The illustration of quantum federated learning process is shown in Fig.~\ref{qai}.

 In designing a variational quantum classifier, the first step is to encode classical data into a quantum state. To efficiently simulate quantum circuits, we employed amplitude encoding, which maps the normalized classical \(N\)-dimensional input data (\(N = 2^n\)) into the amplitudes of an \(n\)-qubit quantum state \(\ket{\psi}\), represented as $
    \ket{\psi} = \frac{1}{\parallel x \parallel} \sum_{j=1}^{N} x_j \ket{j}
$.

Following amplitude encoding, we apply a variational quantum circuit with a limited depth (\textit{l}) to the feature state $\ket{\psi}$, which is composed of single-qubit rotations (RY and RX), followed by a linear arrangement of control NOT gates.  These rotations are parameterized and a classical optimizer is used to update these parameters during training, striving to minimize a predefined loss function. Afterward, compute the circuit's expectation value, which leads to the final result of the classifier. An overview of the variational quantum classifier is shown in Fig.~\ref{vqcfig}.

\subsection{Fisher information-driven optimization}

We introduce the QFedFisher algorithm, an efficient approach for collaborative training across clients, utilizing quantum circuit and Fisher information to enhance model performance while ensuring data privacy. 

\begin{figure}[!ht]
	\centering
\includegraphics[scale=0.65]{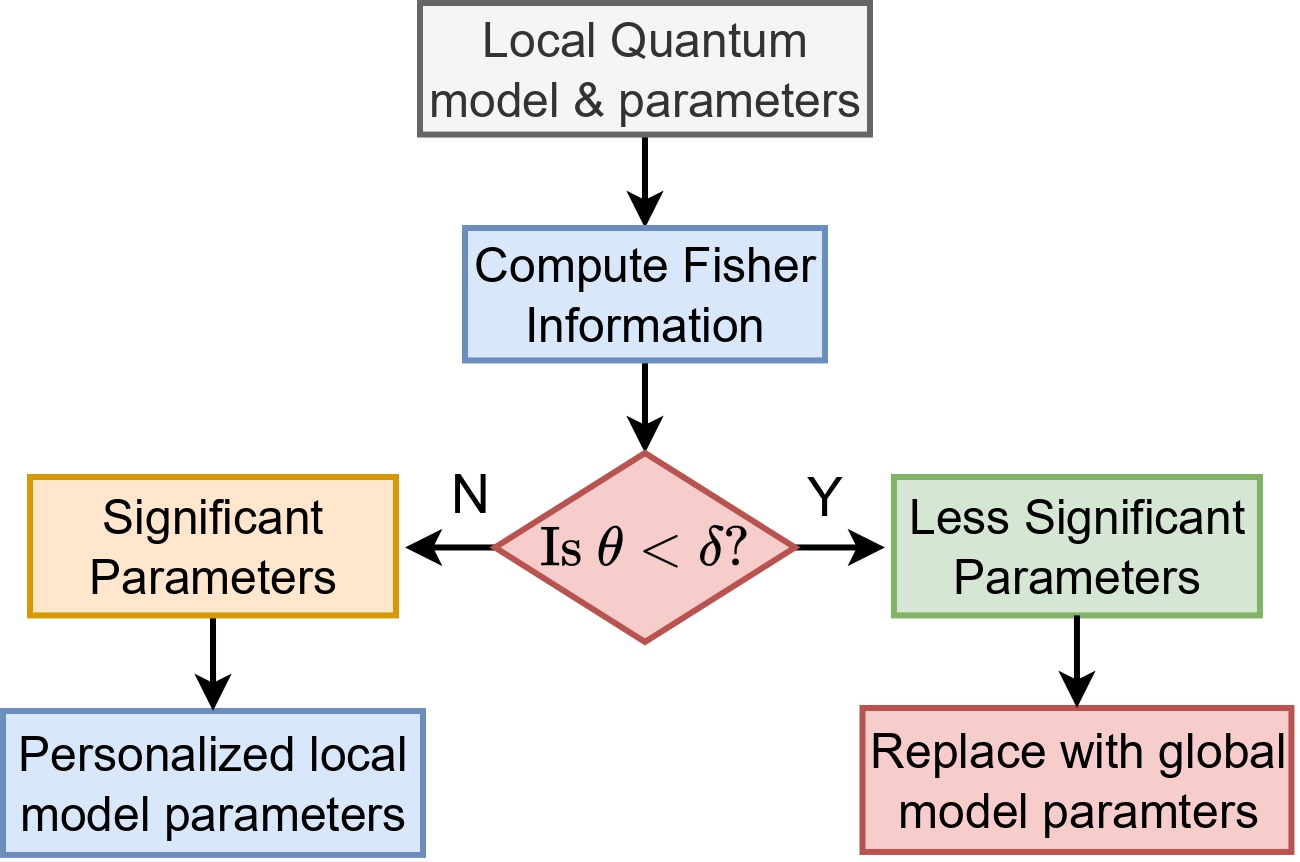}
\caption{\textbf{An illustration of determining the significant parameters using Fisher information}. Each client computes the Fisher information, and the local quantum model parameters ($\theta$) that are less significant (i.e., those below the Fisher threshold $\delta$) are replaced with the global model parameters, while the remaining parameters are retained. The clients then proceed with local training. }
\label{fed}
\end{figure}

Suppose there are multiple clients (\textit{C}) and each client ($i \in C$) has its own dataset containing $D_i$ samples. Initially, a variational quantum classifier (VQC) is employed for local training. For client $i\in C$, the vector of trainable parameters ($\theta$) is represented by $\overrightarrow{\theta}_i=(\theta_1, \theta_2, ..., \theta_{N-1}, \theta_N)^{\intercal}$. During the local client update, a cross-entropy loss function ($\mathcal{L}$) and the ADAM optimizer are used to update the local client quantum circuit parameters. The parameters $\overrightarrow{\theta_i^{r-1}}$ are carried over from the previous communication round \textit{r}. Using this setup, we compute the  Fisher information vector  $\mathcal{F}(\theta_{ij})$ of each client participating in round,  which serves as a reliable approximation of the diagonal of the true Fisher information matrix, providing a parameter-specific measure of sensitivity for each parameter $j$ in $\mathbf{\theta}_{r-1}^i$ as:
\begin{equation}
    \mathcal{F}(\theta_{ij}) = \left( \frac{\partial \log \mathcal{L}(\theta_i, \mathcal{D}_i)}{\partial \theta_{ij}} \right)^2
\end{equation}
Before sending the Fisher information of clients along with their model parameters, the Fisher information matrix is normalized by applying a layer-wise min-max operation. The distinction between significant and less significant parameters based on Fisher information, using a Fisher threshold ($\delta$), and substitution with global model parameters is illustrated in Fig~\ref{fed}.

After completing local training on the client side, the vector of trainable parameters $\overrightarrow{\theta}_i$ and their corresponding Fisher information matrices $\mathcal{F}(\theta_{ij}$) are transmitted to the global server, while the local data remains securely stored at each client. The global model is responsible for coordinating updates from each client and performing the following three steps, after which it distributes the updated parameters to the clients.

\begin{itemize}
    \item Compute weighted average:
   \begin{equation}
        \theta_\text{avg} = \sum_{i=1}^C p_i \theta_i^r
   \end{equation}
    where \(p_i\) is the weight or importance of the \(i\)-th client’s contribution to the global model, which is determined based on the size of the client’s dataset. Typically, \(p_i\) is normalized so that \(\sum_{i=1}^C p_i = 1\),  and \(\theta_i^r\) represents the model parameters of the \(i\)-th client after training in the \(r\)-th round.

    \item Computer Fisher-average gradients and update:
    \begin{equation}G_s = \sum_{i=1}^C \mathcal{F}_{ij} \cdot \theta_i^r, \quad F_s = \sum_{i=1}^C\mathcal{F}_{ij},  \quad \theta_s^r = \frac{G_s}{F_s} \end{equation}

    where \( G_s \) represents the weighted sum of the model parameters \(\theta_i^r\) from all clients with Fisher information \(\mathcal{F}_{ij}\). \( F_s \) is the total sum of the Fisher information matrices \(\mathcal{F}_{ij}\) across all clients.  Finally, \(\theta_s^r\) is the updated global model parameter.

    \item Finally, determine the less significant parameters based on Fisher information and substitute:
 \begin{equation}
     \mathcal{I} = \{ j \mid F_{s} < \delta \}, \quad \theta_{s, j}^r = \theta_{\text{avg}, j} \quad \forall j \in \mathcal{I}.
 \end{equation}
where \(\mathcal{I}\) is the set of indices \(j\) for which the value of \(F_s\) is smaller than a Fisher threshold \(\delta\).  The substitution is performed for the less significant parameters based on the condition \(F_s < \delta\).

\end{itemize}

\section{Experiments}
We conducted extensive experiments on two distinct datasets to assess the performance of the proposed method alongside comparison methods. We tested the accuracy of the proposed method using the ADNI \cite{22} and MNIST \cite{23} and datasets, where data is partitioned in a non-independently and non-identically partitioned (non-IID) manner across clients.

\subsection{Datasets, comparison methods and implementation details}
\begin{figure*}[!ht]
	\centering
\includegraphics[scale=0.55]{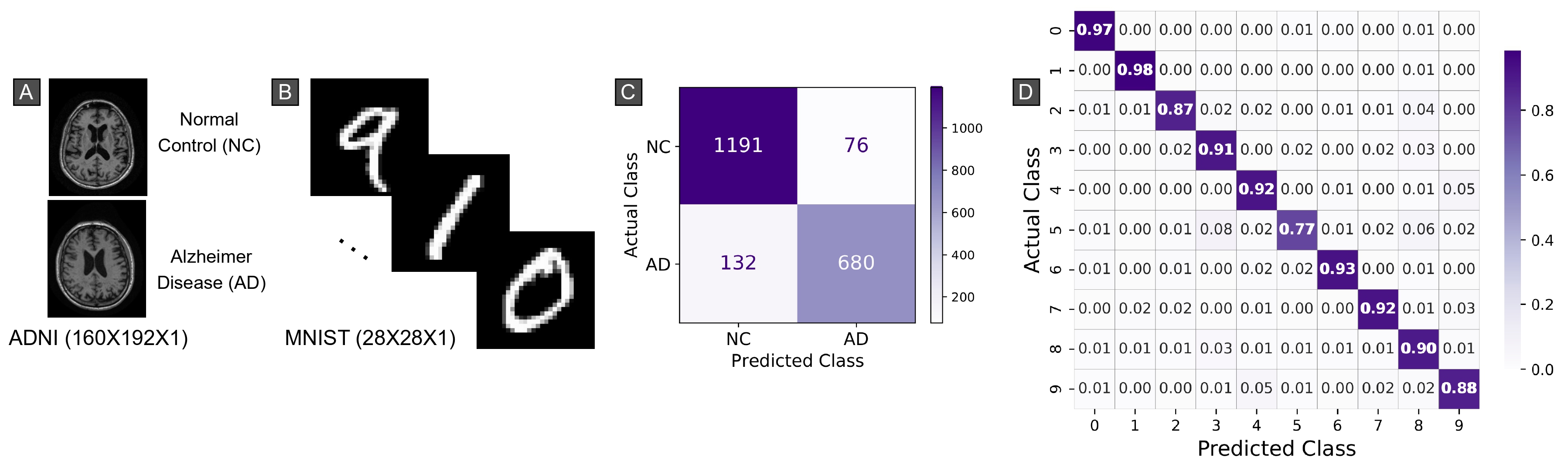}
\caption{\textbf{Sample images and confusion matrices of the quantum federated global model using the QFedFisher algorithm on the testing set} (a) An example of MRI scans of normal cognitive function (NC) and Alzheimer’s disease (AD) (b) SA sample of handwritten MNIST images. (c) Binary classification on the ADNI dataset, distinguishing between normal cognitive function and Alzheimer's disease. (d) Multi-class confusion matrix for the MNIST dataset.}
 \label{confusion}
\end{figure*}

Our method is evaluated on two distinct classification tasks, reflecting real-world non-IID scenarios in federated learning. For binary classification, the ADNI dataset, containing 10,231 MRI scans (3,676 from Alzheimer’s disease (AD) patients and 6,555 from individuals with normal cognition (NC)), is used to differentiate AD from NC. For multi-class classification, the MNIST dataset serves as a benchmark for digit recognition. Sample images from both datasets are shown in Fig.~\ref{confusion}(a-b). We compare our method's performance with state-of-the-art approaches, including QFedAvg and QFedAdam, using testing accuracy of the global model as the primary evaluation metric. Each client trains its local quantum model, and accuracy is assessed on a testing dataset.

We set the number of clients to 100 for the MNIST dataset and 10 for the ADNI dataset. In each round, 5\% of MNIST clients are randomly selected, while all ADNI clients participate. Weights are initialized using Kaiming initialization, and the quantum circuit includes 60 parameterized layers. Data is distributed using a Dirichlet distribution, with each client receiving 500 samples. The Fisher threshold is fixed at 0.01. Each client trains locally for 1 epoch (E=1) with a batch size of 32 (B=32), using the ADAM optimizer with a learning rate of 1e-3. The number of communication rounds is set to 300 for MNIST and 100 for ADNI.

\begin{figure*}[!ht]
	\centering
\includegraphics[scale=0.5]{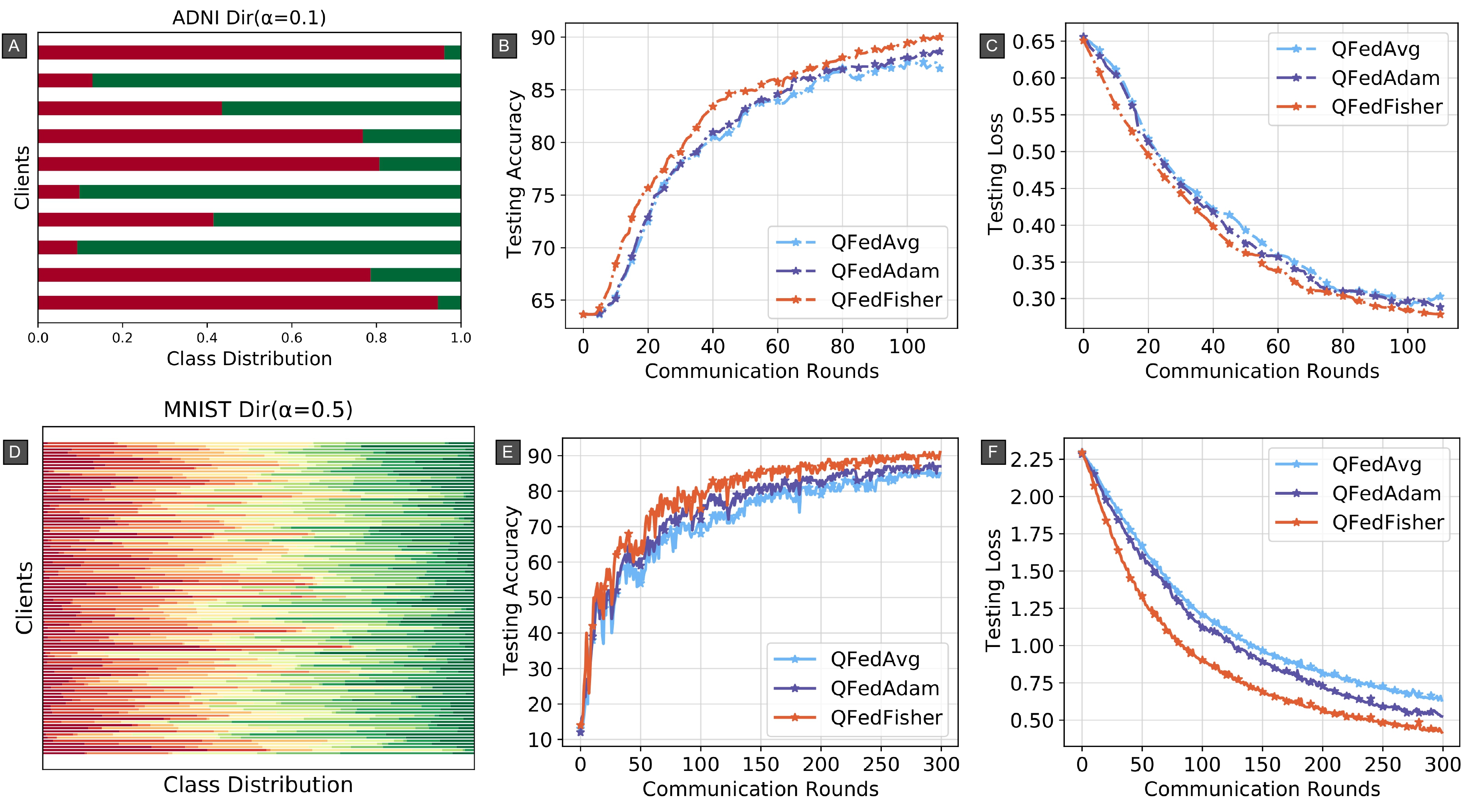}
\caption{\textbf{Performance of quantum federated learning methods on different datasets}. (a)  The unequal distribution of ADNI dataset among 10 clients using Dir($\alpha$=0.1). (b-c) The test accuracy and loss curves of three QFL methods (QFedAvg, QFedAdam, and QFedFisher) on the ADNI dataset, distinguishing between normal and Alzheimer's disease. (d)  The distribution of MNIST (10 colors) dataset using Dir($\alpha$=0.5) and 100 clients (100 rows) on y-axis. (e-f) The performance of global models using three QFL methods on the MNIST dataset, evaluated over 300 fixed communication rounds. }
 \label{main_results}
\end{figure*}

\begin{figure}[!ht]
	\centering
\includegraphics[scale=0.55]{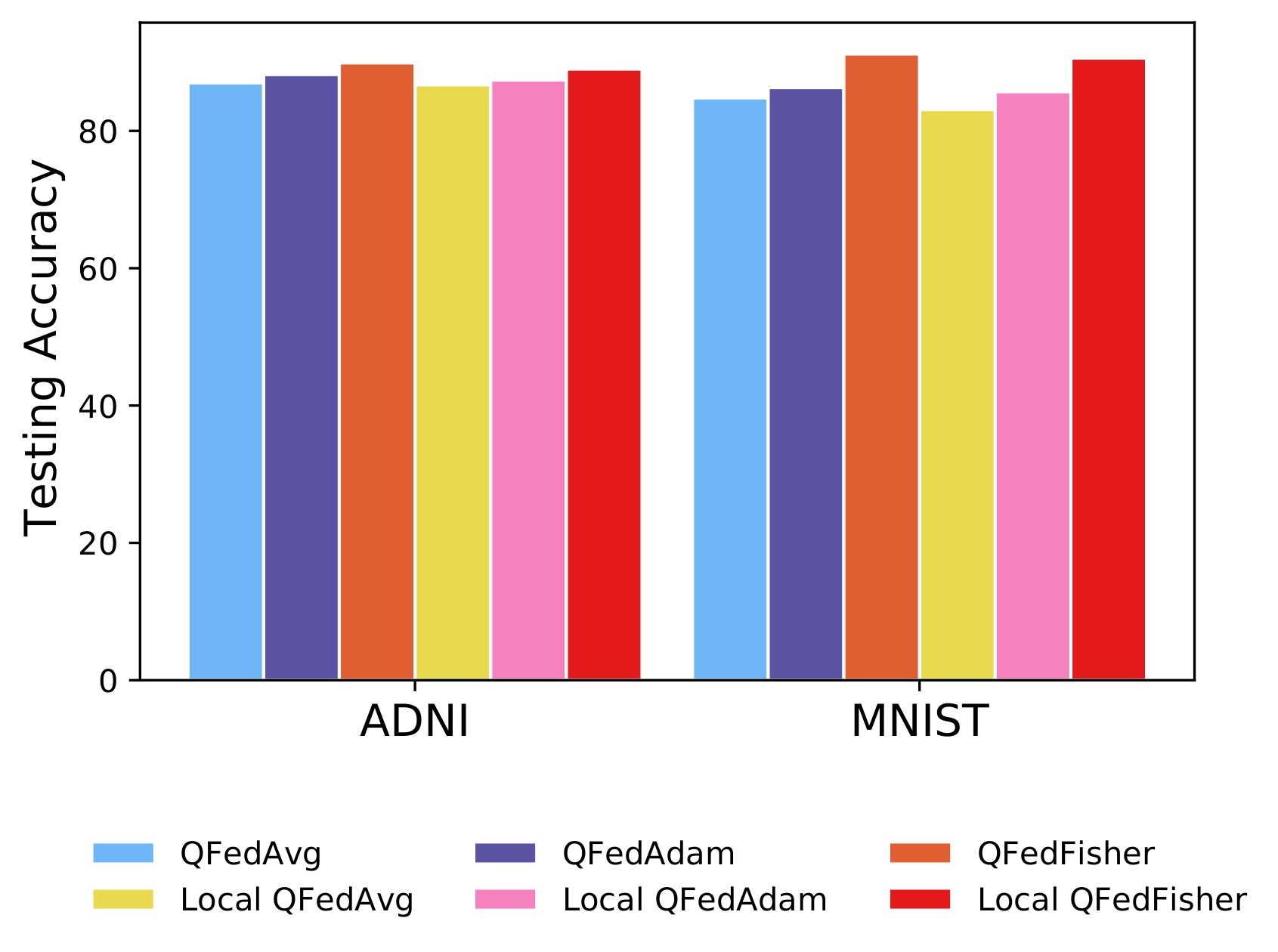}
\caption{\textbf{Comparing the testing accuracy of quantum federated algorithms}.
Also shows the average testing accuracy of their respective local clients, highlighting the consistency between global and local performance of QFedFisher algorithm.}
 \label{bar}
\end{figure}

\section{Results and discussion}
In this section, we demonstrate the effectiveness of the proposed algorithm by comparing its performance with other existing methods to highlight its strengths.

\begin{table}[!ht]
\centering
\caption{Performance (Testing Accuracy and Loss) of different quantum federated models on ADNI and MNIST datasets}
\setlength{\tabcolsep}{5mm} 
\begin{tabular}{@{}l@{\hspace{4mm}}c@{\hspace{3mm}}c@{\hspace{3mm}}c@{\hspace{4mm}}c@{}}
\toprule
Dataset & Models & Accuracy & Loss & Time (secs)  \\
\midrule
\multirow{3}{*}{ADNI} & QFedAvg &   87.0 & 0.302 & 2213.2\\
& QFedAdam &   88.2 & 0.297 & 2396.1 + 8.2\% \\
& QFedFisher &   89.9 & 0.278 & 2481.4 + 12.1\% \\
\midrule
\multirow{3}{*}{MNIST} &  QFedAvg &  84.8 & 0.632 & 974.7  \\
 &QFedAdam &   86.3 & 0.525 & 1098.5 + 12.6\%\\
& QFedFisher &   91.2 & 0.418 & 1128.1 + 15.7\%\\
\bottomrule
\end{tabular}
\label{tab:performance}
\end{table}

\subsection{Model performance on the ADNI dataset}
To evaluate the effectiveness of the proposed algorithm, we analyze the non-IID distribution of the ADNI dataset for binary classification across 10 clients over a fixed 100 communication rounds, as illustrated in Fig~\ref{main_results}(a). Axial 2D slices are extracted from 3D T1-weighted MRI brain images, with the original size of 160$\times$192 reduced to 2048 components using principal component analysis (PCA). This reduction preserved 99\% of the variability while maintaining the key features of the original image. A federated quantum neural network employs an amplitude encoding block to map the 2048 components of an image into 11 qubits. A final measurement is then taken from the last qubit to classify the image into one of two categories: normal or Alzheimer's disease.

For the ADNI dataset, all clients participate in the aggregation process during each communication round. Fig~\ref{main_results}(b-c) presents the results of 100 rounds of aggregation using QFL algorithms on the ADNI dataset. The QFedFisher method demonstrates significant and consistent improvements in both accuracy and convergence speed compared to the QFedAvg and QFedAdam methods. Our proposed method, leveraging the Fisher information from each client's local model, significantly improves the performance of the global model and accurately distinguishes between normal MRI scans and those associated with Alzheimer's disease, despite the uneven data distribution among clients. Fig~\ref{confusion}(c) displays the confusion matrix for the ADNI MRI-scan test set under the QFedFisher approach. The positive predictive value reached 89.9\%, while the negative predictive value was 90.0\%.
Table~\ref{tab:performance} presents the performance of the algorithms on the binary classification ADNI dataset. QFedFisher consistently surpasses the quantum averaging and ADAM baselines, demonstrating its ability to handle data heterogeneity in QFL settings and achieve superior testing accuracy with fewer communication rounds. The QFedFisher method requires less than 12.1\% of the total computational cost of QFedAvg on the client side in our QFL setup for the ADNI dataset, where all 10 clients participate in each round.

\subsection{Model performance on the MNIST dataset}
Next, we evaluate the performance of a variational quantum circuit for a multi-classification task in FL settings. The 28$\times$28 grayscale images from the MNIST dataset are padded with zeroes and encoded into 10 qubits. After encoding, the resulting quantum state is passed through multiple layers of parametrized (trainable) single-qubit rotations (RY and RX) and linearly arranged entangling CNOT gates. Finally, the Z expectation values of the 10 qubits are measured to classify the images into the 10 classes of the MNIST dataset. 

We distribute the dataset among 100 clients using the Dirichlet distribution with a concentration parameter $\alpha=0.5$ and randomly select 5\% of the clients to participate in each round, as shown in Fig~\ref{main_results}(d). All QFL methods are implemented for a fixed 300 communication rounds, and the performance of the global quantum federated model is evaluated on a testing set during each round. Fig~\ref{main_results}(e-f) shows the testing results of the VQC using three different aggregation methods.

We prioritize preserving the most important parameters, which are above the threshold ($\delta$>0.01), as they have a significant impact on the performance of the global model. We find that all quantum federated methods demonstrate smoother convergence and strong generalization capabilities. The global QFedFisher model stands out by significantly surpassing QFedAvg and QFedAdam, achieving higher accuracy and faster convergence. The confusion matrix for the MNIST test set using QFedFisher is shown in Fig.~\ref{confusion}(d). Table~\ref{tab:performance} summarizes the results obtained by the algorithms on the MNIST dataset, along with the total time taken for each communication round on the client side, including the evaluation of performance on the test set. The QFL method utilizing Fisher information demonstrates significant improvement over the baselines, while incurring less than 15.7\% of the total computational cost of QFedAvg at the client side in our QFL setup. The comparison of the testing accuracy of quantum federated algorithms and their local clients is shown in Fig~\ref{bar}.

Calculating the Fisher information for a quantum circuit adds a computational cost, typically less than 15\% of the total time required for the QFedAvg method. This cost depends on factors such as the number of participating clients, the complexity of the quantum circuit (number of qubits and depth), and the specific quantum operations used. Despite the added time, it remains feasible for most practical QFL applications, where the optimization and collaborative learning benefits of incorporating Fisher information outweigh the marginal increase in computation time.

\section{Conclusion}
Integrating Fisher information into quantum federated learning introduces a principled way to optimize client models by leveraging the intrinsic geometry of the parameter space in quantum systems. Parameters that carry more information are better able to represent knowledge, making them essential for accurate model predictions. In this paper, we present quantum federated learning with layer-wise Fisher information (QFedFisher) of the quantum circuit, demonstrating the ability to significantly enhance the effectiveness and convergence of quantum federated learning methods, leading to superior performance on heterogeneous client datasets. Furthermore, it effectively addresses challenges posed by client heterogeneity, where data distributions across clients are significantly non-IID. Fisher Information adjusts updates at both the client and global levels, ensuring that the aggregated model benefits from balanced contributions, even in the presence of data disparity. 
Experimental results on the ADNI and MNIST datasets show that quantum federated learning, utilizing layer-wise Fisher Information of quantum circuits, exhibits greater robustness and achieves higher testing accuracy. This method surpasses the performance of quantum federated averaging and quantum federated Adam within a fixed number of communication rounds. 
In the future, we plan to extend our work on quantum federated learning with Fisher Information by incorporating privacy-preserving techniques. Since Fisher Information helps identify the most important parameters at the client level, these techniques can provide privacy guarantees by ensuring that sensitive parameters are protected during model aggregation. 

\section*{Acknowledgments}
We would like to acknowledge the partial internal support from the ECE Department at NC State University and the U.S. Department of Energy (DOE) (Office of Basic Energy Sciences), under Award No. DE-SC0019215.

	\bibliography{mainfile}
	
\end{document}